\pdfoutput=1
\documentclass[11pt]{article}

\usepackage[]{EACL2023}

\usepackage{times}
\usepackage{latexsym}

\usepackage[T1]{fontenc}
\usepackage[utf8]{inputenc}

\usepackage{microtype}

\usepackage{inconsolata}

\usepackage{booktabs}
\usepackage{hyperref}
\usepackage{multirow}
\usepackage{subcaption}
\usepackage{graphicx}
\usepackage{amsmath}
\usepackage{mathtools}

\title{Comparing Intrinsic Gender Bias Evaluation Measures \\without using Human Annotated Examples}

\author{Masahiro Kaneko$^{1}$ \quad
        Danushka Bollegala$^{2,3}$\Thanks{Danushka Bollegala holds concurrent appointments as a Professor at University of Liverpool and as an Amazon Scholar. This paper describes work performed at the University of Liverpool and is not associated with Amazon.} \quad
        Naoaki Okazaki$^{1}$ \\
        $^1$Tokyo Institute of Technology \quad
        $^2$University of Liverpool \quad
        $^3$Amazon \\
        {\tt masahiro.kaneko@nlp.c.titech.ac.jp} \\
        {\tt danushka@liverpool.ac.uk} \quad
        {\tt okazaki@c.titech.ac.jp}
}

\begin{document}
\maketitle
\begin{abstract}
Numerous types of social biases have been identified in pre-trained language models (PLMs), and various intrinsic bias evaluation measures have been proposed for quantifying those social biases.
Prior works have relied on human annotated examples to compare existing intrinsic bias evaluation measures.
However, this approach is not easily adaptable to different languages nor amenable to large scale evaluations due to the costs and difficulties when recruiting human annotators.
To overcome this limitation, we propose a method to compare intrinsic gender bias evaluation measures without relying on human-annotated examples.
Specifically, we create multiple \emph{bias-controlled} versions of PLMs using varying amounts of male vs. female gendered sentences, mined automatically from an unannotated corpus using gender-related word lists. 
Next, each bias-controlled PLM is evaluated using an intrinsic bias evaluation measure, and the rank correlation between the computed bias scores and the gender proportions used to fine-tune the PLMs is computed.
Experiments on multiple corpora and PLMs repeatedly show that the correlations reported by our proposed method that does not require human annotated examples are comparable to those computed using human annotated examples in prior work.
\end{abstract}

\section{Introduction}

Pre-trained language models (PLMs) trained on large datasets have reported impressive performance improvements in various NLP tasks~\cite{devlin-etal-2019-bert,lan2019albert} greatly.
However, these PLMs also demonstrate significantly worrying levels of social biases~\cite{NIPS2016_a486cd07,kurita-etal-2019-measuring}.
To address this issue, numerous intrinsic bias evaluation measures for PLMs have been proposed~\cite{nangia-etal-2020-crows,bold_2021,nadeem-etal-2021-stereoset,kaneko2022unmasking,zhou-etal-2022-sense}, which are also used for comparing debiasing methods for PLMs~\cite{webster2020measuring,kaneko-bollegala-2021-debiasing,schick-etal-2021-self}.

Existing bias evaluation methods use different criteria such as pseudo likelihood~\cite{kaneko2022unmasking}, cosine similarity~\cite{WEAT,may-etal-2019-measuring}, inner-product~\cite{ethayarajh-etal-2019-understanding} etc.
Moreover, current bias evaluation methods require manually-annotated datasets containing stereotypical and antistereotypical examples that express different types of social biases ~\cite{nangia-etal-2020-crows,nadeem-etal-2021-stereoset}.
Therefore, we consider that it is important to compare the differences in existing bias evaluation measures proposed for PLMs~\cite{orgad-belinkov-2022-choose,Dev2021OnMO,kaneko-etal-2022-debiasing} to understand their relative strengths and weaknesses. 

\begin{figure}[t]
  \centering
  \includegraphics[width=0.49\textwidth]{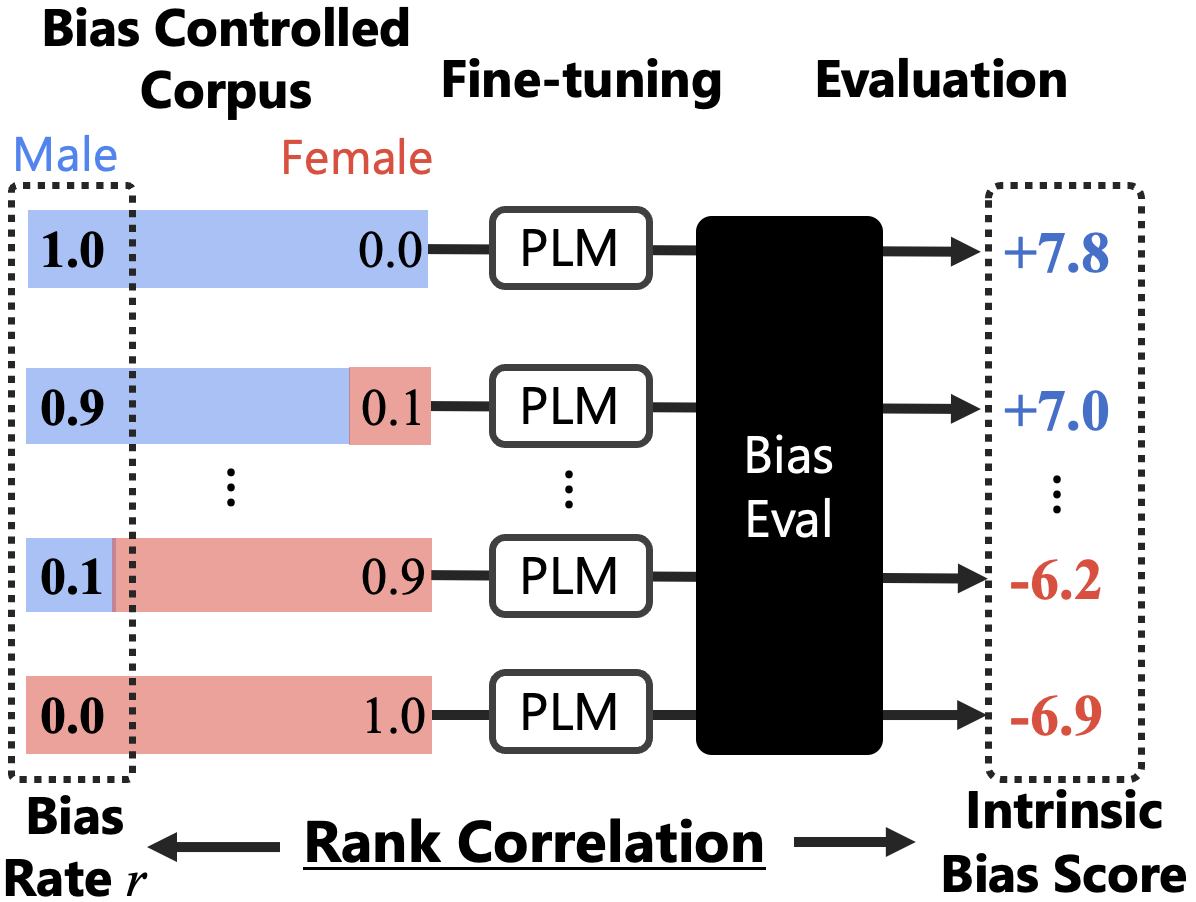}
  \caption{Overview of our proposed method.
  We first create \emph{bias-controlled} PLMs by fine-tuning a PLM on male and female gendered sentences that are automatically mined from unannotated corpora. 
  Next, we measure the rank correlation between the scores reported by an intrinsic bias evaluation measure and the male/female bias rates ($r$) used to fine-tune the PLMs.}
  \label{fig:method}
\end{figure}

To objectively compare the existing bias evaluation measures, \citet{kaneko2022unmasking} calculated the rank correlation between the number of human annotators who labelled an example to be stereotypically biased towards a protected attribute in Crowds-Pairs (CP), and the bias score for that example returned by an intrinsic bias evaluation measure~\cite{nangia-etal-2020-crows,nadeem-etal-2021-stereoset}.
However, due to the costs and difficulties in recruiting human annotators, this approach cannot be easily adapted to different languages, accommodate large-scale evaluations, or compare evaluation metrics that do not use human-annotated data.

We propose a method to compare intrinsic bias evaluation measures without using human annotated examples.
\autoref{fig:method} outlines the intuition behind our proposed method.
First, we train \emph{bias-controlled} versions of PLMs obtained via fine-tuning a PLM on male and female gendered sentences, automatically mined from an unannotated corpus using a gender-related word list.
We define \emph{rate of bias} ($r$) as the ratio between male and female gendered sentences in a training sample used to fine-tune a PLM.
A PLM fine-tuned mostly on male sentences is likely to generate sentences containing mostly male words, while a PLM fine-tuned on female sentences is likely to generate sentences containing mostly female words~\cite{kaneko2022unmasking,kaneko-etal-2022-gender}.
Therefore, an accurate intrinsic bias evaluation measure is expected to return a score indicating a bias towards the male gender for a male bias-controlled PLM, while it is expected to return a score indicating a bias towards the female gender for a female bias-controlled PLM.
We then compute the rank correlation between (a) the rate of biases in the bias-controlled PLMs, and (b) the bias scores returned by an intrinsic evaluation measure for the corresponding PLMs, as a measure of accuracy of the bias evaluation measure.

Our experiments with multiple corpora and PLMs show that the correlations reported by our proposed method, which does not require human annotated examples, are comparable to those computed using human annotated examples in previous studies.
Furthermore, by examining the output probabilities of the PLM, we verify that the proposed method, which fine-tunes bias-controlled PLMs with varying amounts of male vs. female sentences, is indeed able to control biases associated with male and female gender directions.

\section{Bias-controlled Fine-Tuning}

The imbalance of gender words in the training data affects the gender bias of a PLM fine-tuned using that data~\cite{kaneko2022unmasking,kaneko-etal-2022-gender}.
%Therefore, we adjust the number of gender words in the training data to control the model's bias.
Using this fact, we propose a method to learn \emph{bias-controlled} versions of PLMs that express different levels of known gender biases.
Let us first assume that we are given a list of female gender related words $\mathcal{V}_f$ (e.g. \textit{she, woman, female}), and a separate list of male gender related words $\mathcal{V}_m$ (e.g. \textit{he, man, male}).
Next, we select sentences that contain either at least one of female or male words from an unannotated set of sentences $\mathcal{D}$.
Sentences that contain both male and female words are excluded here.
Let us denote the set of sentences extracted for a female or a male word $w$ by $\Phi(w)$.
Moreover, let $\mathcal{D}_f = \bigcup_{w \in \mathcal{V}_f} \Phi(w)$ and $\mathcal{D}_m = \bigcup_{w \in \mathcal{V}_m} \Phi(w)$ be the sets of sentences containing respectively female and male words.
We appropriately downsample $\mathcal{D}_f$ and $\mathcal{D}_m$ to have equal numbers of sentences $N$ (i.e. $|\mathcal{D}_f| = |\mathcal{D}_m| = N$).

Next, we create training datasets $\mathcal{D}_r$ by varying the rate of bias, $r$ ($\in [0,1]$), by randomly sampling a subset $\mathcal{S}_r(\mathcal{D}_m)$ of $Nr$ sentences from $\mathcal{D}_f$ and a subset $\mathcal{S}_{1-r}(\mathcal{D}_f)$ of $N(1-r)$ sentences from $\mathcal{D}_m$ such that $\mathcal{D}_r = \mathcal{S}_r(\mathcal{D}_m) \cup \mathcal{S}_{1-r}(\mathcal{D}_f)$.
%Training data $\mathcal{D}_{r}$ sampled at $r$ is created by $S_{r}(\mathcal{D}_m) + S_{1 - r}(\mathcal{D}_f)$.
%Here, sampling function $S_{r}$ samples sentences by $N \times r$.
When $r = 0$, $\mathcal{D}_r$ consists of only female sentences (i.e. $\mathcal{D}_r \subseteq \mathcal{D}_f$), and when $r = 1$, it consists of only male sentences (i.e. $\mathcal{D}_r \subseteq \mathcal{D}_m$).
To obtain multiple bias-controlled PLMs at different levels of gender biases, we fine-tune a given PLM on different datasets, $D_{r}$, sampled with different values of $r$.
We use a given intrinsic bias evaluation measure to separately evaluate each bias-controlled PLM.
Finally, we measure the agreement between the bias scores reported by the intrinsic bias evaluation measure under consideration and the corresponding rates of biases of those PLMs using Pearson's rank correlation coefficient.

\section{Experiments}
\label{sec:exp}

\subsection{Settings}
\label{sec:exp:settings}

% 追加のモデル学習の話

In our experiments, we used the female words \textit{she, woman, female, her, wife, mother, girl, sister, daughter, girlfriend} as $\mathcal{V}_f$, and male words \textit{he, man, male, him, his, husband, father, boy, brother, son, boyfriend} as $\mathcal{V}_m$.
We sampled 2M sentences each representing male and female genders from News crawl 2021 corpus (\textbf{news})\footnote{\url{https://data.statmt.org/news-crawl/en/}} and BookCorpus~\cite{zhu2015aligning} (\textbf{books}) for training bias-controlled PLMs and a separate 100K sentences as development data.
We used BERT\footnote{\url{https://huggingface.co/bert-base-uncased}}~\cite{devlin-etal-2019-bert} and ALBERT\footnote{\url{https://huggingface.co/albert-base-v2}}~\cite{lan2019albert} as the PLMs.
We fine-tune PLMs with masked language model learning.
We use publicly available Transformer library\footnote{\url{https://github.com/huggingface/transformers/tree/v4.22.2}} to fine-tuning PLMs, and all hyperparameters are set to their default values.
We trained 11 bias-controlled PLMs for $r$ in $\{0.0, 0.1, 0.2, 0.3, 0.4, 0.5, 0.6, 0.7, 0.8, 0.9, 1.0\}$ on four Tesla V100 GPUs.

\subsection{Intrinsic Bias Evaluation Measures}
\label{sec:intrinsic}

We compare five previously proposed intrinsic gender bias evaluation measures in this paper: Template-Based Score~\cite[\textbf{TBS};][]{kurita-etal-2019-measuring}, StereoSet Score~\cite[\textbf{SSS};][]{nadeem-etal-2021-stereoset}, CrowS-Pairs Score~\cite[\textbf{CPS};][]{nangia-etal-2020-crows}, All Unmasked Likelihood~\cite[\textbf{AUL};][]{kaneko2022unmasking}, and AUL with Attention weights~\cite[\textbf{AULA};][]{kaneko2022unmasking}.
Further details of these measures are given in the Appendix.

Note that TBS uses templates for evaluation and cannot be used with human-annotated stereotypical/anti-stereotypical sentences.
On the other hand, SSS, CPS, AUL, and AULA all require human-annotated sentences that express social biases.

\subsection{Comparing Intrinsic Gender Bias Evaluation Measures}

\begin{table}[!t]
\centering
\small
\begin{tabular}{lrrrrrr}
\toprule
 \multirow{2}{*}{Measure} & \multicolumn{3}{c}{BERT} & \multicolumn{3}{c}{ALBERT} \\
 \cmidrule(r){2-4}
 \cmidrule(r){5-7}
& news & book & HA & news & book & HA  \\
\midrule
TBS     & 0.14 & 0.09 & -    &  0.25 & 0.14 & - \\
SSS     & 0.22 & 0.22 & 0.45 &  0.31 & 0.22 & 0.53 \\
CPS     & 0.30 & 0.27 & 0.57 &  0.37 & 0.22 & 0.48 \\
AUL     & 0.37 & 0.32 & 0.68 &  0.55 & 0.36 & 0.56 \\
AULA    & 0.42 & 0.34 & 0.71 &  0.60 & 0.42 & 0.57 \\
\bottomrule
\end{tabular}
\caption{Peason correlation between biased PLM order and each bias scores. News and book represent the corpus used for biasing, respectively. HA is AUC value of method using human annotation~\cite{kaneko-bollegala-2021-debiasing}.}
\label{tbl:meta_eval_mlm}
\end{table}

We compare the proposed method and \citet{kaneko2022unmasking}'s method using CP dataset, which has human annotations, and show the effectiveness of the proposed method.
In addition, we will use several PLMs and corpora to analyze the trends of the proposed method.
\autoref{tbl:meta_eval_mlm} shows the correlation results of the proposed method for TBS, SSS, CPS, AUL, and AULA when fine-tuning BERT and ALBERT on news or book corpora, respectively.
HA is the AUC value of the \citet{kaneko2022unmasking}'s method using human annotations.
Since TBS uses templates, it cannot be evaluated using HA.

For BERT, the proposed method induces the same order among measures (i.e. AULA $>$ AUL $>$ CPS $>$ SSS) as done by HA in both news and book.
For ALBERT, only the rankings of SSS and CPS differ between the proposed method and HA.
These results show that the proposed method and the existing method that use human annotations rank the intrinsic gender bias evaluation measures in almost the same order.\footnote{Because of the different methods of measuring correlations, it is not possible to compare the magnitude of values between the proposed and existing methods.}
It can be seen that the values of the correlation coefficients vary depending on the PLM and corpus.
For example, ALBERT has a maximum correlation of 0.60, while BERT has a maximum correlation of only 0.42.

A major limitation of human annotation-based evaluation is that it cannot be used to compare TBS that does not human annotated examples against other intrinsic bias evaluation measures.
However, our proposed method does \emph{not} have this limitation and can be used to compare TBS against other bias evaluation measures.
As it can be seen from \autoref{tbl:meta_eval_mlm}, TBS consistently reports the lowest correlations, indicating that it is not an accurate intrinsic gender bias evaluation measure.
This finding agrees with \citet{kaneko-etal-2022-debiasing}, who highlighted the inadequacy of templates as a method for evaluating social biases.

\subsection{Bias-controlled PLMs}

\begin{figure}[t!]
    \centering
    \includegraphics[width=0.5\textwidth]{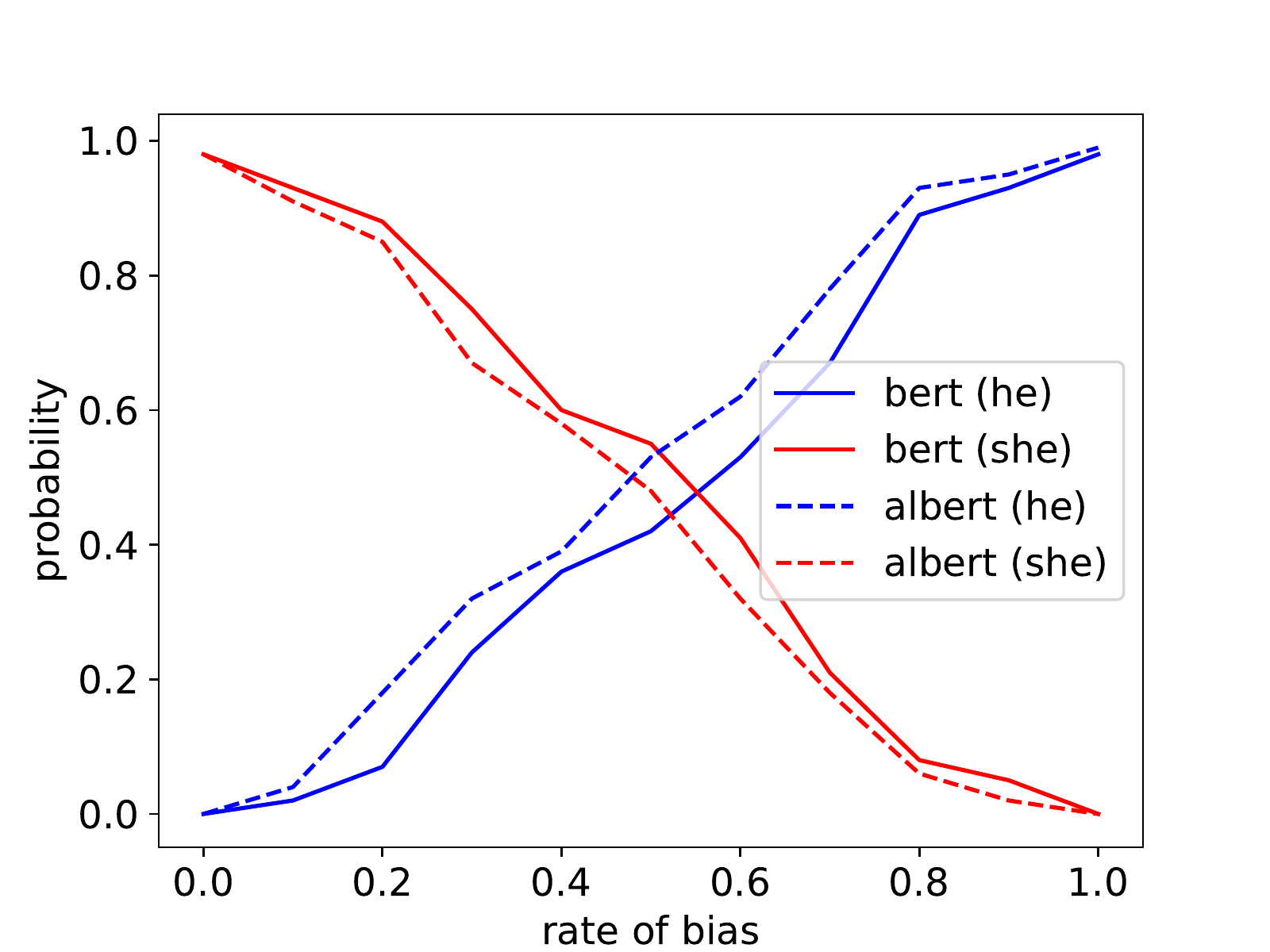}
    \caption{Average output probabilities for ``\textit{[MASK] is a/an [Occupation]}'' produced by the bias-controlled BERT and ALBERT PLMs fine-tuned with different $r$ on the news dataset.}
    \label{fig:prob}
%\vspace{-5mm}
\end{figure}

To verify that the proposed method can indeed control the bias of a PLM, we investigate the variation of the output probabilities of the PLMs fine-tuned with different $r$.
Specifically, we investigate the output probabilities of masked \textit{he} and \textit{she} in the input text ``\textit{[MASK] is a/an [Occupation].}'' for the bias-controlled PLMs.
For \textit{[Occupation]}, we use gender- and stereotype-neutral occupational words\footnote{\url{https://github.com/tolga-b/debiaswe}} (e.g. \textit{writer}, \textit{musician}) from the word list created by \citet{NIPS2016_a486cd07}.
%As the value of $r$ increases, the average output probability of \textit{he} increases, and if the average output probability of \textit{she} decreases, the proposed method can control the bias of the PLM.
When $r$ increases, a PLM will be fine-tuned with increasing amounts of male sentences.
Therefore, if the average probability of \emph{he} increases with $r$, it would imply that the PLMs are correctly bias-controlled by the proposed method.

\autoref{fig:prob} shows that the average output probabilities of \textit{he} and \textit{she} when $r$ is incremented in step size of 0.1.
%Orig is the result of BERT without biasing.
When $r = 1$ the PLM predicts \textit{he} with fairly high probability and when $r = 0$ the PLM predicts \textit{she} with fairly high probability.
Furthermore, when $r = 0.5$, the probability of \textit{he} and \textit{she} is almost 0.5.
Original BERT (without fine-tuning) returns 0.48 and 0.28, respectively for \emph{he} and \emph{she}, while the corresponding probabilities returned by ALBERT are respectively 0.64 and 0.22.
Both the original BERT and ALBERT predict relatively larger output probabilities for \emph{he}, indicating that they are male-biased, without performing any bias-control.
From these results, it can be seen that the output probabilities of \textit{he} and \textit{she} fluctuate according to $r$, and the proposed method can control the bias of the PLM.
On the other hand, when $r$ is less than 0.2 or greater than 0.8, the output probabilities of \textit{she} and \textit{he} are greater than the proportion in the data set, respectively.
Therefore, finer increments of $r$ may make it difficult to control bias more finely when $r$ is small or large.

\begin{figure}[t!]
    \centering
    \begin{subfigure}{0.44\textwidth}
    \includegraphics[width=\textwidth]{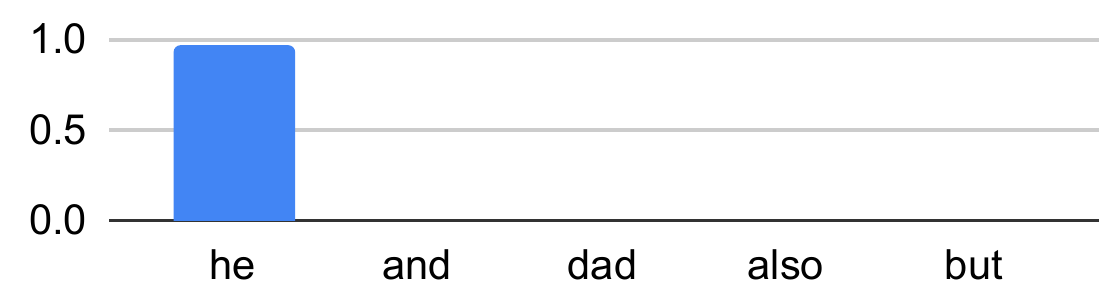}
          \caption{$r = 1.0$}
    \end{subfigure}
    \begin{subfigure}{0.44\textwidth}
    \includegraphics[width=\textwidth]{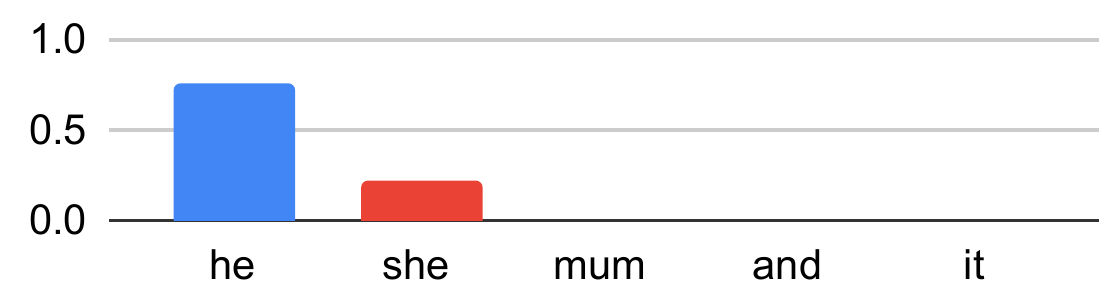}
          \caption{$r = 0.7$}
    \end{subfigure}
    \begin{subfigure}{0.44\textwidth}
    \includegraphics[width=\textwidth]{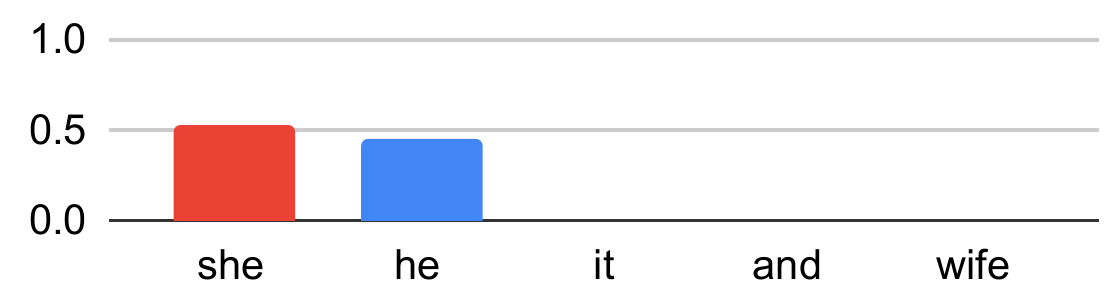}
          \caption{$r = 0.5$}
    \end{subfigure}
    \begin{subfigure}{0.44\textwidth}
    \includegraphics[width=\textwidth]{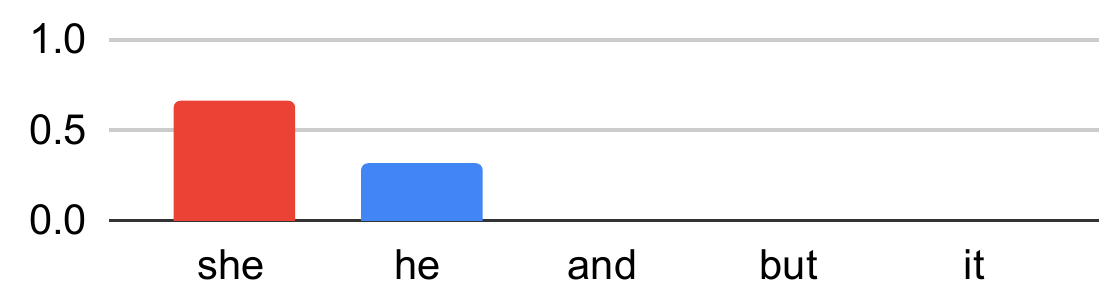}
          \caption{$r = 0.3$}
    \end{subfigure}
    \begin{subfigure}{0.44\textwidth}
    \includegraphics[width=\textwidth]{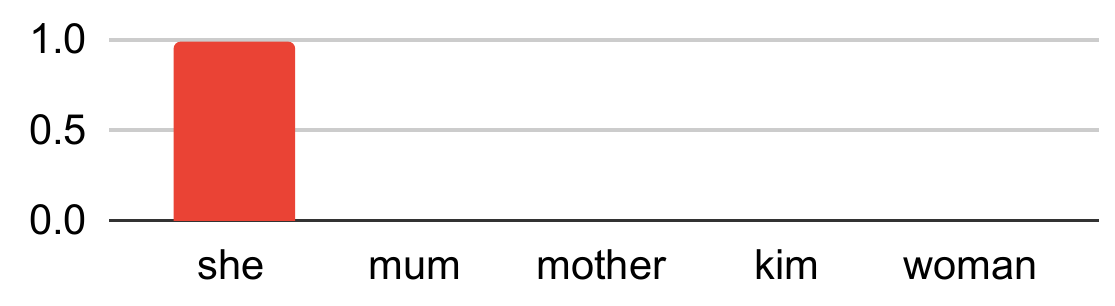}
          \caption{$r = 0.0$}
    \end{subfigure}
    \begin{subfigure}{0.44\textwidth}
    \includegraphics[width=\textwidth]{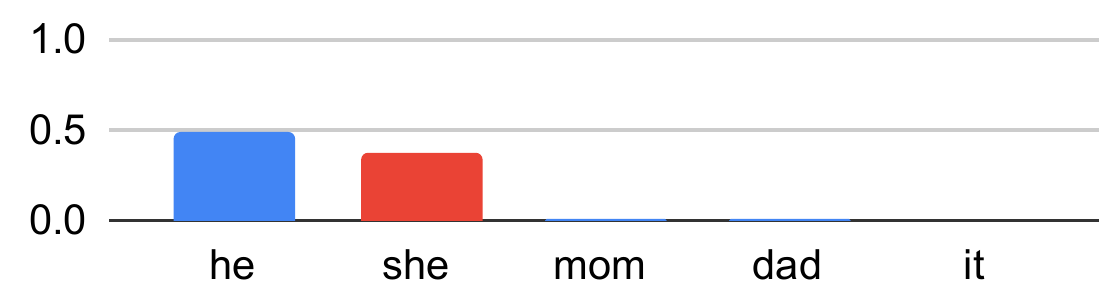}
          \caption{original (without fine-tuning)}
    \end{subfigure}
\caption{Top 5 words with BERT output probability for ``\textit{[MASK] doesn't have time for family due to work obligations.}''. Blue and red represent masculine and feminine words, respectively.}
\label{fig:example}
%\vspace{-5mm}
\end{figure}

To illustrate how bias-controlled PLMs produced by the proposed method for different rates of biases ($r$) predict the probabilities of gender pronouns, we consider the masked sentence 
``\textit{[MASK] doesn't have time for the family due to work obligations.}'' selected from the CP dataset.
Here, \textit{He} and \textit{She} are unmodified tokens.
\autoref{fig:example} shows the probabilities of the tokens predicted for the [MASK] by the different bias-controlled PLMs.
We see that the original BERT model predicts both \emph{he} and \emph{she} with approximately equal probabilities.
However, when $r$ is gradually increased from 0 to 1, we see that the probability of \emph{he} increases, while that of \emph{she} decreases, demonstrating that the proposed method correctly learns bias-controlled PLMs.

\section{Conclusion}

We proposed a method to compare intrinsic gender bias evaluation measures using an unannotated corpus and gender-related word lists.
%We fine-tune various bias-controlled PLMs using different proportions of sentences with different genders as male or female.
%A rank correlation is then calculated between the bias scores predicted by the intrinsic bias evaluation measure and the gender ratios used to fine-tune the PLM.
%According to our experiments, the correlations reported by the proposed method, which does not require human-annotated examples, are consistently comparable to previous studies using human-annotated examples.
Experiments show that the correlations computed by the proposed method for existing bias evaluation measures agrees with the prior evaluations conducted using human annotations.

\section{Limitations}
\label{sec:limitations}

In this paper, we limited our investigation to English PLMs.
However, as reported in a lot of previous work, social biases are language independent and omnipresent in PLMs trained for many languages~\cite{kaneko-etal-2022-gender,lewis2020gender,liang-etal-2020-monolingual,zhao-etal-2020-gender}.
We plan to extend this study to cover non-English PLMs in the future.

According to existing research, PLMs encode many different types of social biases such as racial and religious biases in addition to gender-related biases~\cite{kiritchenko-mohammad-2018-examining,ravfogel-etal-2020-null}.
On the other hand, in this paper, we focused on only gender bias.
Extending the proposed method to handle other types of social biases beyond gender bias is beyond the scope of the current short paper and is deferred to future work.

Furthermore, discriminatory bias is learned in word embeddings as well as PLMs~\cite{NIPS2016_a486cd07,brunet2019understanding,kaneko-bollegala-2019-gender,kaneko-bollegala-2020-autoencoding,kaneko-bollegala-2021-dictionary,kaneko2022gender}.
Therefore, it may be possible to make it applicable to word embeddings as well.

\section{Ethical Considerations}
\label{sec:ethics}

Our goal in this paper was to compare the previously proposed and widely-used intrinsic bias evaluation measures of gender bias in pre-trained PLMs.
Although we used a broad range of existing datasets that are annotated for social biases, we did not annotate nor release new datasets as part of this research.
Moreover, we fine-tune a large number of bias-controlled PLMs for evaluation purposes that demonstrates varying levels of gender biases.
However, these PLMs are not supposed to be used in downstream tasks other than for evaluation purposes.

Even with the highly correlated bias evaluation measure in our proposed method, the bias of the PLM may not be sufficiently evaluated.
Therefore, we consider that it important to select intrinsic gender bias evaluation measures carefully and not purely based on correlation coefficients computed by the proposed method alone.

There are various discussions on how to define social bias in PLMs~\cite{blodgett-etal-2021-stereotyping}.
Since the proposed method can use any method as the bias-controlled fine-tuning of the PLMs, the bias-controlled fine-tuning can be selected according to the definition of social bias.

\section*{Acknowledgements}

This paper is based on results obtained from a project, JPNP18002, commissioned by the New Energy and Industrial Technology Development Organization (NEDO).

\bibliography{custom}
\bibliographystyle{acl_natbib}

\appendix
\section{Intrinsic Bias Evaluation Measures}

\paragraph{TBS}
\citet{kurita-etal-2019-measuring} proposed template-based bias evaluation measure.
The log-odds of the likelihood of a template sentence masked with a gender word (e.g. \textit{``[MASK] is a programmer''}) and the likelihood of a gender word masked with an occupation word (e.g. \textit{``[MASK] is a [MASK]''}) are calculated for male and female words, respectively.
TBA then calculates the difference between them as the bias score.

\paragraph{SSS}
SSS~\cite{nadeem-etal-2021-stereoset} uses stereotypical and anti-stereotypical sentence pairs (e.g. \textit{``She is a nurse''} and \textit{``He is a nurse''}) to evaluate bias in PLMs.
Calculate the likelihood of masked modified tokens (e.g. \textit{She, He}) given unmasked unmodified tokens (e.g. \textit{is, a, nurse}) for each stereotypical and anti-stereotypical sentence.
The bias score is calculated by dividing the number of sentences for which the total likelihood is higher for stereotypical sentences compared to anti-stereotypical sentences by the total number of data.

\paragraph{CPS}
CPS~\cite{nangia-etal-2020-crows} also uses stereotypical and anti-stereotypical sentence pairs.
On the other hand, calculate the likelihood of masked unmodified tokens given unmasked modified tokens for each stereotypical and anti-stereotypical sentence.
The bias score is calculated by dividing the number of sentences for which the total likelihood is higher for stereotypical sentences compared to anti-stereotypical sentences by the total number of data.
As with SSS, the bias score is calculated using the sum of the likelihoods of the stereotyped and anti-stereotyped sentences.

\paragraph{AUL and AULA}
AUL and AULA~\cite{kaneko2022unmasking} also uses stereotypical and anti-stereotypical sentence pairs, but they calculate the likelihood of unmasked unmodified tokens and modified tokens for each stereotypical and anti-stereotypical sentence.
As with SSS and CPS, the bias score is calculated using the sum of the likelihoods of the stereotyped and anti-stereotyped sentences.
AULA calculates the likelihood of the entire sentence by weighting and averaging with the attention weights to prioritize the likelihood of important words.

\end{document}